\documentclass[10pt,twocolumn,letterpaper]{article}

\usepackage{wacv}
\usepackage{times}
\usepackage{epsfig}
\usepackage{graphicx}
\usepackage{amsmath}
\usepackage{amssymb}
\usepackage{hyperref}
\hypersetup{
	colorlinks = true, 
	urlcolor = magenta, 
	linkcolor = red, 
	citecolor = blue}

\usepackage{enumitem}
\usepackage{romannum}
\usepackage{booktabs}
\usepackage{amsmath}

\usepackage{array}
\newcolumntype{P}[1]{>{\centering\arraybackslash}p{#1}}

\def\wacvPaperID{68} 

\wacvfinalcopy 

\ifwacvfinal
\fi

\begin{document}
	\title{Pixel-Level Bijective Matching for Video Object Segmentation}
	\author{Suhwan Cho\quad Heansung Lee\quad Minjung Kim\quad Sungjun Jang\quad Sangyoun Lee\vspace{0.5cm}\\
		Yonsei University\\}
	\maketitle
	
	\pagenumbering{gobble} 
	
	\begin{abstract}
		Semi-supervised video object segmentation (VOS) aims to track the designated objects present in the initial frame of a video at the pixel level. To fully exploit the appearance information of an object, pixel-level feature matching is widely used in VOS. Conventional feature matching runs in a surjective manner, i.e., only the best matches from the query frame to the reference frame are considered. Each location in the query frame refers to the optimal location in the reference frame regardless of how often each reference frame location is referenced. This works well in most cases and is robust against rapid appearance variations, but may cause critical errors when the query frame contains background distractors that look similar to the target object. To mitigate this concern, we introduce a bijective matching mechanism to find the best matches from the query frame to the reference frame and vice versa. Before finding the best matches for the query frame pixels, the optimal matches for the reference frame pixels are first considered to prevent each reference frame pixel from being overly referenced. As this mechanism operates in a strict manner, i.e., pixels are connected if and only if they are the sure matches for each other, it can effectively eliminate background distractors. In addition, we propose a mask embedding module to improve the existing mask propagation method. By embedding multiple historic masks with coordinate information, it can effectively capture the position information of a target object. Code and models are available at \url{https://github.com/suhwan-cho/BMVOS}.

	\end{abstract}
	
	\begin{figure}[t]
		\centering
		\includegraphics[width=1.0\linewidth]{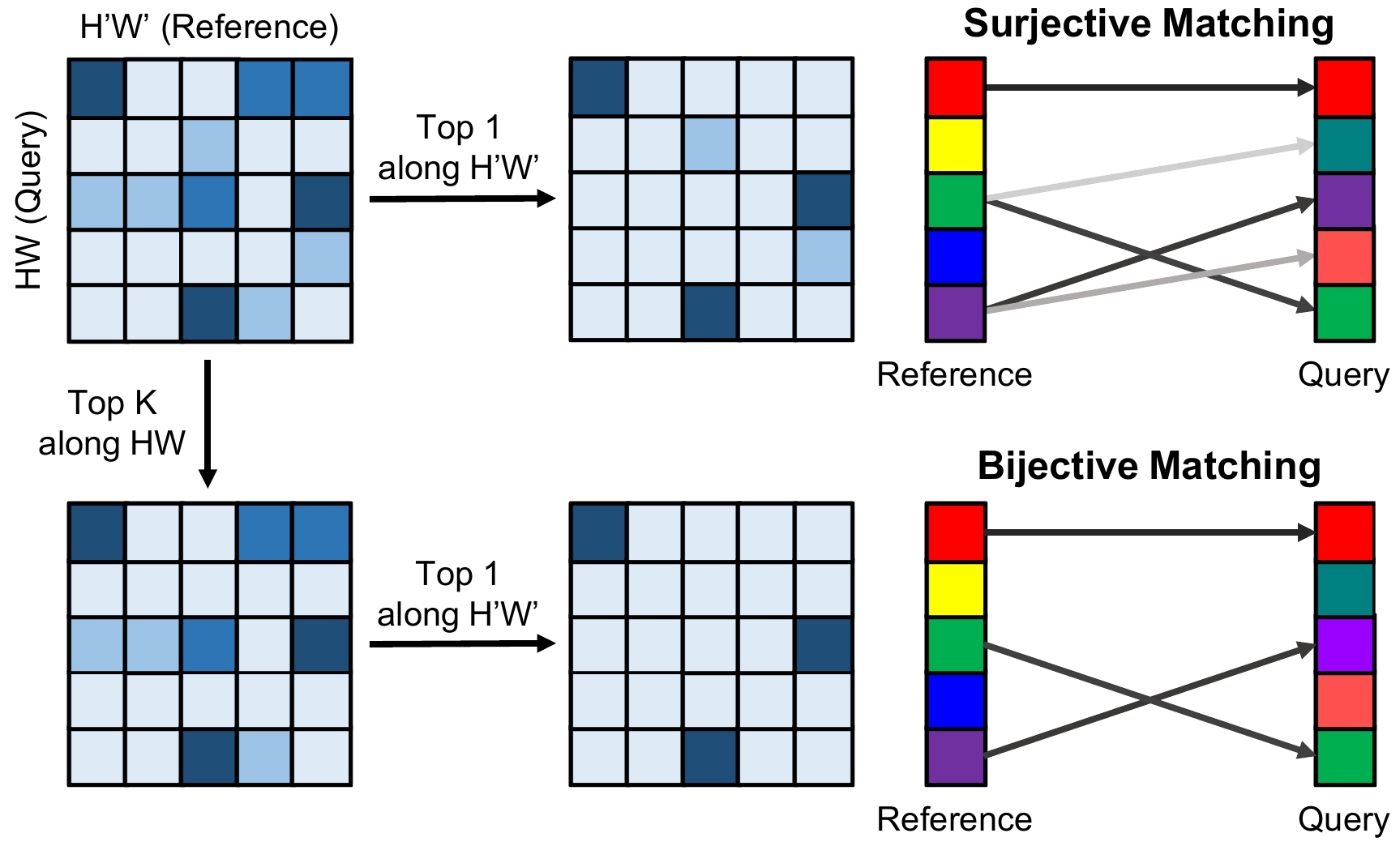}
		\caption{Comparison of a conventional surjective matching mechanism and our proposed bijective matching mechanism by visualizing the similarity map generation process and information transfer flow. We assume that the reference frame contains H'W' pixels and query frame contains HW pixels. Surjective matching transfers information from the reference frame to the query frame considering only the best matches for the query frame, while bijective matching considers the best matches for both the reference frame and the query frame.}
		\label{figure1}
	\end{figure}

	\section{Introduction}
	In recent years, video object segmentation (VOS) has attracted extensive attention because of its applicability to diverse fields in computer vision. It is also widely used in real-world applications including video editing, autonomous driving, video surveillance, and robotics. VOS can be divided into different subcategories, such as semi-supervised VOS, unsupervised VOS, weakly-supervised VOS, interactive VOS, and referring VOS, depending on the type of target object guidance. In this study, we deal with semi-supervised VOS, which aims to track and segment a designated target object in a video using the initial frame ground truth segmentation mask of that object.

	To leverage the appearance information of the target object, feature matching that compares the extracted features is widely adopted. As VOS is a pixel-level task, the feature matching is generally performed for every single pixel features to capture the detailed information. The features extracted from the query frame are compared to those from the reference frame at the pixel level. Then, based on the similarity scores from the comparison, the information of the reference frame is transferred to the query frame.

	Conventional VOS solutions~\cite{PLM, RANet, STM, CFBI} employ a surjective matching mechanism. That is, the query frame pixels select the best-matching pixels in the reference frame and transfer the information from those pixels without any consideration of reference frame options. This approach is robust against large appearance changes and works well in most cases, but may cause critical errors when background distractions are spread on the query frame. As there is no limitation to the number of reference frame pixels being referenced, background distractors in the query frame will get high foreground scores and can disrupt the prediction. To mitigate this issue, we introduce a bijective matching mechanism to find the best matches from the query frame to the reference frame and also vice versa. In Figure~\ref{figure1}, we visualize the comparison of surjective matching and bijective matching. Unlike surjective matching, the pixels in bijective matching are connected for information transfer if and only if they are the best-matching pixels for each other. As information transfer is performed much more strictly compared to surjective matching, it is effective when the reference frame and query frame are visually similar and contain multiple distractions in the background.

	Although feature matching is effective at capturing the appearance information of a target object, it is prone to distractions. To compensate for this, some works~\cite{RGMP, FEELVOS, A-GAME} use the position information of an object to take advantage of the property of a video that an object usually occupies similar positions in consecutive frames. Despite its soundness, the use of position information is still limited under a simple mask propagation method, such as feeding the downsampled previous frame mask into a decoder. To better utilize the position information of a target object, we propose a mask embedding module. By considering multiple historic masks simultaneously, the future object positions can be better predicted. To help the network perceive the spatial location, coordinate information is also fed into the mask embedding module.

	We validate our proposed method on public benchmark datasets for VOS, i.e., the DAVIS 2016~\cite{DAVIS2016}, DAVIS 2017~\cite{DAVIS2017}, and YouTube-VOS 2018~\cite{YTVOS} datasets. Our proposed bijective matching mechanism and mask embedding module are demonstrably effective on all datasets. In addition, compared to existing state-of-the-art methods, our method achieves comparable accuracy with a much higher inference speed. We believe our method has the potential to be widely used in various computer vision applications.

	Our main contributions can be summarized as follows:
	\begin{itemize}[leftmargin=0.2in]
		\item We introduce a bijective matching mechanism to find the best matches from the query frame to the reference frame and vice versa. As information is transferred in a strict manner, background distractions in the query frame can be effectively removed. 
		
		\item We propose a mask embedding module that embeds multiple historic masks with coordinate information to better predict the future object positions.
		
		\item Our proposed method achieves favorable accuracy compared to existing state-of-the-art methods while maintaining much faster inference speed.
	\end{itemize}

	\begin{figure*}[t]
		\centering
		\includegraphics[width=0.92\linewidth]{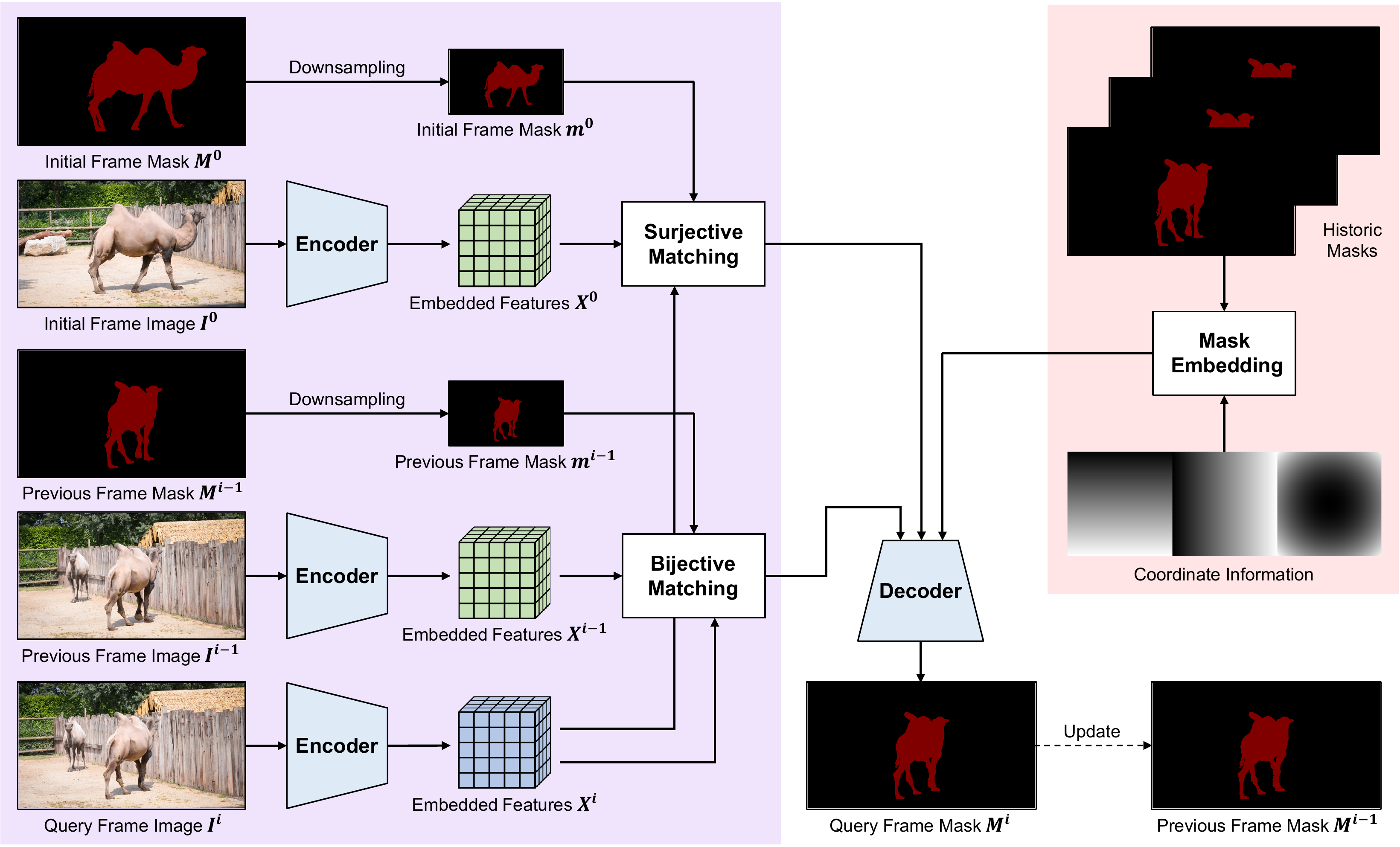}
		\caption{Architecture of our proposed method. Surjective matching is used for global matching (initial frame as a reference frame), and bijective matching is used for local matching (previous frame as a reference frame). To utilize the position information of a target object, multiple historic frames are embedded with coordinate information. The output scores from feature matching modules and features from mask embedding module are then fed into a decoder for query frame mask prediction.}
		\label{figure2}
	\end{figure*}

	\section{Related Work}
	\noindent\textbf{Feature similarity matching.} Semi-supervised VOS is generally based on feature matching, which compares the embedded features of reference frames against a query frame. To capture the information that lies in a local area, the feature matching for VOS is performed at the pixel level. VideoMatch~\cite{VideoMatch} builds foreground and background templates using the initial frame information, and generates the similarity scores using a soft matching layer. Extending from VideoMatch, FEELVOS~\cite{FEELVOS} uses the previous frame's information as well as the initial frame when generating the template, and utilizes the output of feature matching as an internal guidance of the network. To increase the usefulness of the generated similarity map, RANet~\cite{RANet} proposes a ranking attention module that ranks and selects conformable feature maps as per their importance. In order to fully utilize the information extracted from all past frames, TVOS~\cite{TVOS} proposes a transductive approach and STM~\cite{STM} proposes to use memory networks. Extending from STM, EGMN~\cite{EGMN} exploits an episodic memory network where frames are stored as nodes. Memory-based approaches are effective at fully utilizing the accumulated information but suffer from increasing memory size over time. To overcome this limitation, GC~\cite{GC} proposes a global context module to store the information with constant memory size and computation. Similarly, AFB-URR~\cite{AFB-URR} introduces an adaptive feature bank update scheme to dynamically absorb new features and discard obsolete features for efficient memory management. 
	
	\vspace{1mm}
	\noindent\textbf{Background distraction.} There are several existing works that focus on relieving background distractors. FEELVOS applies a window for local matching with a previous frame to exclude the distractors by restricting the area when identifying the most similar pixel in the template. CFBI~\cite{CFBI} also applies windows for local matching but uses multiple windows to deal with various degrees of motion effectively. The most similar work to our work is KMN~\cite{KMN}. It proposes a kernelized memory read operation to relieve distractions by exploiting the locality of VOS. During the memory read process, the amount of information transferred from memory to query is adjusted by a 2D Gaussian kernel that is generated by an argmax operation along a query frame. However, as this mechanism is based on the spatial distance, it is not able to erase the background distractors that are entangled with a foreground object. In that case, it is still surjective as all query frame pixels can refer to the reference frame pixels. Unlike these works, our proposed matching mechanism is completely bijective as query frame pixels can refer to reference frame pixels only if they are selected by reference frame pixels. As it is quite strict (not all query frame pixels are matched), it can prevent background distractors from bringing foreground information.

	\vspace{1mm}
	\noindent\textbf{Mask propagation.} To exploit the property of a video that consecutive frames are strongly interdependent, mask propagation is popularly used in semi-supervised VOS. RGMP~\cite{RGMP} embeds the previous frame segmentation mask with the query frame image to extract features of the query frame while considering the position of the target object at the previous frame. RANet, FEELVOS, and CFBI feeds a downsampled segmentation mask into a decoder to provide a coarse guidance. CRVOS~\cite{CRVOS} even solves semi-supervised VOS using only mask propagation without feature matching. In this study, we extend existing mask propagation methods by extracting features from multiple historic masks to utilize their temporal variations.

	\section{Approach}
	The goal of semi-supervised VOS is to predict the segmentation masks for an entire video sequence using the ground truth segmentation mask given at the initial frame. Our framework can be divided into two portions based on the types of information being used. The portion that utilizes appearance information consists of feature matching modules, while the portion that uses position information is composed of a mask embedding module. The overview of our framework is illustrated in Figure~\ref{figure2}.

	\subsection{Feature Similarity Matching}
	\label{feature matching}
	Like most state-of-the-art approaches~\cite{RMNet, LCM, GIEL, SSTVOS} for VOS, we use pixel-level feature matching that compares the features of every spatial location for tracking an object. The input image and extracted features at frame $i$ are denoted as $I^i \in [0,255]^{3 \times H0 \times W0}$ and $X^i \in \mathbb{R}^{C \times H \times W}$, respectively. The predicted segmentation mask and a downsampled version of it are given by $M^i \in [0,1]^{2 \times H0 \times W0}$ and $m^i \in [0,1]^{2 \times H \times W}$, where the first channel indicates background and the second channel indicates the foreground. As the ground truth segmentation mask of the initial frame is given, every spatial location in $M^0$ can be determined to 0 or 1. Given that frame $i$ is a query frame and frame $k \in [0,i-1]$ is a reference frame, our objective is to generate $M^i$ by using $X^i$, $X^k$, and $m^k$.
	
	\vspace{1mm}
	\noindent\textbf{Surjective matching.} To capture the relationship between the reference frame and the query frame, we first compute the similarity scores based on cosine distance between two pixels in the feature space. If $p$ and $q$ are the single pixel locations in reference frame and query frame features, the similarity score between those pixels can be calculated as
	\begin{align}
	&sim(p, q) = \frac{\mathcal{N}(X^k_p) \cdot \mathcal{N}(X^i_q) + 1} {2}~,
	\label{eq1}
	\end{align}
	where $\mathcal{N}$ and $\cdot$ indicate channel normalization and matrix inner product, respectively. To force the similarity score to be in the range of 0 to 1 for calculation with predicted segmentation mask, we linearly normalize the cosine similarity value. Then, the similarity score matrix, $S \in [0,1]^{H'W' \times HW}$, can be obtained, where $H'W'$ and $HW$ are the spatial size of reference frame and query frame features, respectively. To embody the target object information, reference frame segmentation mask $m^k$ is reshaped, expanded for query frame spatial location, and then multiplied to the similarity score matrix $S$ as
	\begin{align}
	&S_{BG} = S \odot m^k_0\nonumber\\
	&S_{FG} = S \odot m^k_1~,
	\label{eq2}
	\end{align}
	where $\odot$ indicates Hadamard product and $m^k_0$ and $m^k_1$ denote the first and second channels of $m^k$, respectively. Finally, by applying query-wise maximum operation to $S_{BG}$ and $S_{FG}$ respectively, the matching scores $Y \in [0,1]^{H \times W}$ can be obtained for each class. The values in $Y$ indicate the extent to which each pixel location in query frame is similar to each class, i.e., the probabilities of belonging to each class based on the reference frame information.

	If we delve into the above feature matching process, it can be observed that the matching mechanism is surjective. In other words, when connecting reference frame pixels to query frame pixels, only the query frame pixels have the option. All query frame pixels select specific spatial pixel locations in a reference frame and transfer information from those pixels, considering the similarity scores. As the optimal matches from reference frame to query frame are not considered, some reference frame pixels may be referenced multiple times while others are not even referenced. This property makes surjective matching robust against severe scale variations or rapid appearance changes, but also susceptible to background distractions in its nature. In other words, this surjective property of conventional matching mechanism makes the algorithm flexible, but at the same time quite unreliable.

	\begin{figure}[t]
		\centering
		\includegraphics[width=1.0\linewidth]{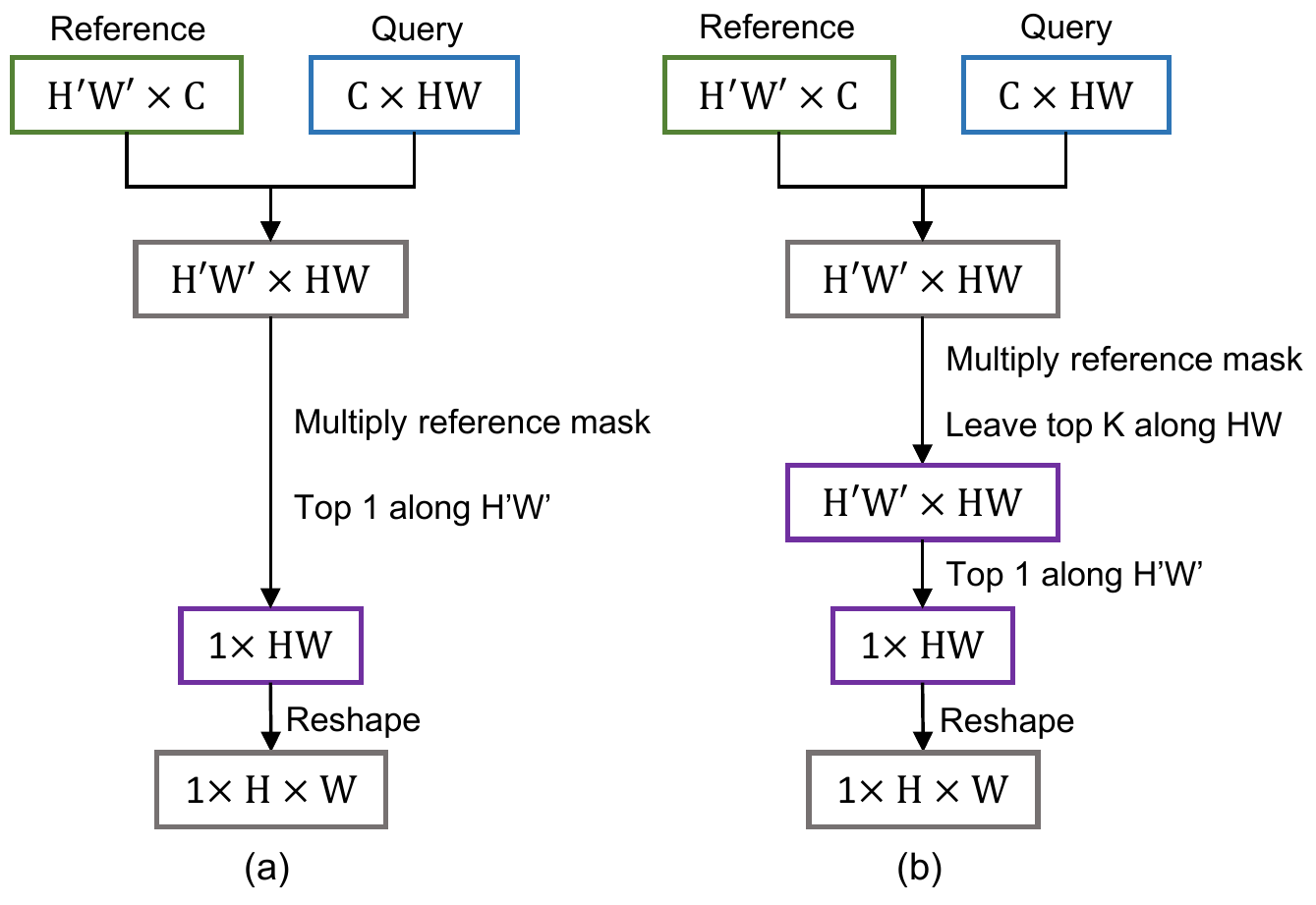}
		\caption{Pipelines of (a) a surjective matching mechanism and (b) a bijective matching mechanism. Bijective matching can be implemented by simply adding a top K operation along the query frame for surjective matching.}
		\label{figure3}
	\end{figure}

	\vspace{1mm}
	\noindent\textbf{Bijective matching.} To mitigate this concern related to surjective matching, we introduce a bijective matching mechanism. We compare the pipelines of two matching methods in Figure~\ref{figure3}. Unlike traditional surjective matching, our bijective matching performs a reference-wise top K operation before finding the best matches from the query frame to the reference frame so as to remove background distractors from consideration. To ensure homogeneity between the connected pixels, only the maximum K scores along query frame are preserved while the others are discarded. The similarity scores for the discarded connections are replaced by the lowest value in each reference frame pixel. If K is set to infinity, there is no discarded connection so the matching is equivalent to the surjective matching. As can be seen, query frame pixels can refer to the reference frame pixels if and only if they are selected by at least one reference frame pixel. The query frame pixels that are not chosen are not able to transfer information from the reference frame. This means bijective matching operates in a stricter manner compared to surjective matching, and this enables safer information transfer between the two frames.

	As described, surjective matching and bijective matching have different advantages that can complement each other. Surjective matching is effective when the reference frame and query frame are visually different but is prone to background distractors. By contrast, bijective matching is effective for transferring information between two frames that are strongly dependent but is not effective for handling visually different frames because of its strictness. To fully exploit their respective advantages, we use surjective matching for global matching with the initial frame and bijective matching for local matching with the previous frame. The matching scores generated from the two different matching methods are fed into the decoder to be used as the appearance information of a target object.

	\subsection{Mask Embedding}
	\label{mask embedding}
	Although feature matching is effective for utilizing appearance information, it is susceptible to visual distractions as it is completely based on visual information. To mitigate this issue, we also use position information to complement the feature matching. As consecutive frames in a video are strongly related, we can infer that the position of a target object will also be similar in consecutive frames. The most common approach to exploit this locality of VOS is to propagate the previous frame’s predicted mask for current frame mask prediction. Generally, this mask propagation is simply performed by feeding a downsampled segmentation mask of the previous frame to a decoder. Through this, the network finds an object around the target location at the previous frame and can exclude background distractors to a certain extent. However, compared to the importance of position information, existing mask propagation methods have considerable scope for improvement.

	\begin{figure}[t]
		\centering
		\includegraphics[width=1.0\linewidth]{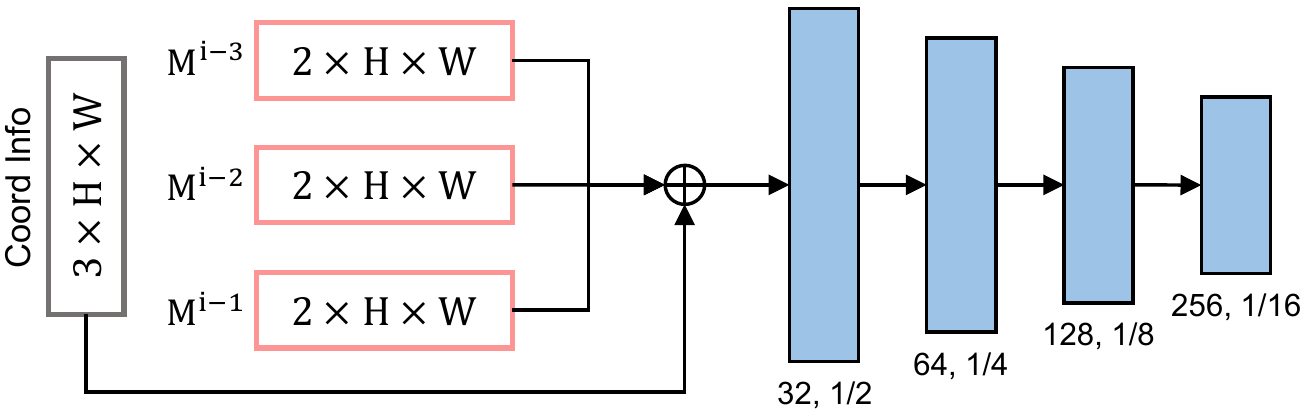}
		\caption{Architecture of our mask embedding module. $\oplus$ indicates concatenation along channel dimension.}
		\label{figure4}
	\end{figure}

	To better exploit position information of a target object, we propose a mask embedding module that extracts effective information from historic target positions. In Figure~\ref{figure4}, we visualize the architecture of our mask embedding module. Unlike existing mask propagation methods~\cite{A-GAME, FEELVOS, RANet} that only use the previous frame mask, we use multiple historic masks to capture the tendencies of position variation. The input of our mask embedding module is defined as
	\begin{align}
	&M^{i-3} \oplus M^{i-2} \oplus M^{i-1}~,
	\label{eq20}
	\end{align}
	where $\oplus$ indicates concatenation along channel dimension. By providing multiple historic masks simultaneously, the position-based coarse prediction for the current frame’s target location can be obtained. In addition, to make the network understand the spatial location, coordinate information~\cite{coord} is also concatenated along the channel dimension. We use three pieces of coordinate information, i.e., height, width, and the distance from the center point. We then apply four convolutional layers with ReLU activation functions~\cite{relu} to embed target position features. The embedded position features are fed into the decoder with the output from feature matching modules.

	\subsection{Implementation Details}
	\noindent\textbf{Encoder.} We use the DenseNet-121~\cite{densenet} that is pre-trained on ImageNet~\cite{imagenet} as our encoder for rich feature representations. As a vanilla version outputs 1/32-sized feature maps before fully-connected layers compared to the input image, it is difficult to capture the details for pixel-level classification. Therefore, we only use the first three blocks to obtain fine-grained feature maps. The highest feature map generated by the encoder has a channel size of 1024 and spatial size of 1/16 compared to the input image.

	\vspace{1mm}
	\noindent\textbf{Decoder.} The goal of a decoder is to merge high-level features from different modules and refine those features using multi-level features generated by an encoder. In our framework, the similarity score maps from feature matching modules and embedded position features from the mask embedding module are fed into a decoder as input. They are then gradually refined and upsampled using multi-level features extracted from the encoder using skip connections. This feature refining process is essential, but simply upsampling high-level features takes excessive computational time. To obtain efficiency, we add channel-reducing deconvolution layers~\cite{deconv} after every skip connection as in CRVOS~\cite{CRVOS}.

	\vspace{1mm}
	\noindent\textbf{Multi-object processing.} To deal with a video with multiple target objects, some early works~\cite{RGMP, STM, KMN, RMNet} independently process each target object one-by-one. In this case, the computational cost increases linearly based on the number of target objects. Unlike theses works, our method shares the image feature embedding, which is the most computational process in the entire pipeline. This makes our method quite efficient for multi-object processing. After predicting the segmentation mask for each target object, we aggregate those masks using soft aggregation as in prior works~\cite{RGMP, A-GAME, STM}.

	\subsection{Network Training}
	\label{network training}
	To overcome the lack of VOS data, existing state-of-the-art solutions~\cite{STM, KMN, CFBI, RMNet} use static image datasets, such as MSRA10K~\cite{MSRA10K}, ECSSD~\cite{ECSSD}, PASCAL-S~\cite{PASCAL-S}, PASCAL-VOC~\cite{PASCAL-VOC}, and COCO~\cite{COCO} datasets. They use them to simulate video samples by augmenting the image samples or use the image samples themselves to obtain better feature representations. Unlike those methods that require complex network training strategies, our model is trained with a simple strategy. Either DAVIS 2017~\cite{DAVIS2017} training set or YouTube-VOS 2018~\cite{YTVOS} training set is used, and the models are tested on the respective datasets. Each training snippet comprises of 10 frames with a spatial size of 384$\times$384 that are randomly cropped from multiple consecutive frames jointly and batch size is set to 8. We randomly select a single foreground object as a target object, and other objects are considered as the background. If there is no target object observed in the initial frame, that training sample is skipped. To prevent the network from being biased towards background attributes, we design each training snippet to contain enough number of target object pixels in the initial frame, similar to CFBI~\cite{CFBI}. We use cross-entropy loss and an Adam optimizer~\cite{adam} with a learning rate of 1e-4 without learning rate decay. During the training stage, all batch normalization layers~\cite{batchnorm} are disabled and the encoder is frozen to preserve rich feature representations. Network training is implemented on a single GeForce RTX 2080 Ti GPU.

	\section{Experiments}
	In this section, we describe the experimental results obtained from this study. The experimental setup, i.e., datasets and evaluation metrics, are described in Section~\ref{setup}.  Quantitative comparison with state-of-the-art methods on public benchmark datasets can be found at Section~\ref{quantitative}. To validate the effectiveness of each component, we perform an extensive ablation study in Section~\ref{ablation}. Our method is abbreviated as BMVOS.

	\subsection{Experimental Setup}
	\label{setup}
	\noindent\textbf{Datasets.} We use DAVIS 2016~\cite{DAVIS2016}, DAVIS 2017~\cite{DAVIS2017}, and YouTube-VOS 2018~\cite{YTVOS} datasets to validate our proposed approach. DAVIS 2016 is the most basic dataset, comprising 30 training videos and 20 validation videos. DAVIS 2017 is an extended version of DAVIS 2016, containing 60 training videos, 30 validation videos, and 30 test-dev videos. YouTube-VOS 2018 is the largest dataset for VOS, containing 3,471 training videos and 474 validation videos. All video sequences of DAVIS 2016 only contains a single target object, while the videos of DAVIS 2017 and YouTube-VOS 2018 may contain multiple target objects.
	
	\vspace{1mm}
	\noindent\textbf{Evaluation metrics.} Generally, intersection-over-union is used to evaluate the segmentation performance in VOS. The accuracy is calculated for the entire region using region accuracy $\mathcal{J}$ and for the object boundaries using contour accuracy $\mathcal{F}$. Usually, the overall accuracy $\mathcal{G}$, which is the average of $\mathcal{J}$ and $\mathcal{F}$, is used as a main metric.

	\begin{table}
		\centering 
		\caption{Quantitative evaluation on the DAVIS 2016 validation set. OL indicates online learning. (+S) indicates the use of static image datasets during the network training.}
		\vspace{2mm}
		\small
		\begin{tabular}{lP{0.58cm}P{0.58cm}P{0.58cm}P{0.58cm}P{0.58cm}}
			\toprule
			Method &OL &fps &$\mathcal{G}_\mathcal{M}$ &$\mathcal{J}_\mathcal{M}$ &$\mathcal{F}_\mathcal{M}$\\
			\midrule
			RMNet~(+S)~\cite{RMNet} & &11.9 &81.5 &80.6 &82.3\\
			FEELVOS~(+S)~\cite{FEELVOS} & &2.22 &81.7 &80.3 &83.1\\
			RGMP~(+S)~\cite{RGMP} & &7.69 &81.8 &81.5 &82.0\\
			DTN~(+S)~\cite{DTN} & &14.3 &83.6 &83.7 &83.5\\
			RANet~(+S)~\cite{RANet} & &30.3 &85.5 &85.5 &85.4\\
			CFBI~(+S)~\cite{CFBI} & &5.56 &86.1 &85.3 &86.9\\
			STM~(+S)~\cite{STM} & &6.25 &86.5 &84.8 &88.1\\
			PReMVOS~(+S)~\cite{PReMVOS} &\checkmark &0.03 &86.8 &84.9 &\textbf{88.6}\\
			KMN~(+S)~\cite{KMN} & &8.33 &\textbf{87.6} &\textbf{87.1} &88.1\\
			\midrule
			RGMP~\cite{RGMP} & &7.69 &68.8 &68.6 &68.9\\
			RANet~\cite{RANet} & &30.3 &- &73.2 &-\\
			FAVOS~\cite{FAVOS} & &0.56 &81.0 &82.4 &79.6\\
			FRTM~\cite{FRTM} &\checkmark &21.9 &81.7 &- &-\\
			\midrule
			\textbf{BMVOS} & &\textbf{45.9} &\textbf{82.2} &\textbf{82.9} &\textbf{81.4}\\
			\bottomrule
		\end{tabular}
		\label{Table:DAVIS16_val}
	\end{table}

	\subsection{Quantitative Results}
	\label{quantitative}
	We quantitatively compare our method with state-of-the-art methods on the DAVIS 2016~\cite{DAVIS2016}, DAVIS 2017~\cite{DAVIS2017}, and YouTube-VOS 2018~\cite{YTVOS} datasets. For a fair comparison, the methods which employ other video training datasets are not reported. The methods are separately compared depending on the use of static image datasets.

	\begin{table}
		\centering 
		\caption{Quantitative evaluation on the DAVIS 2017 validation set. OL indicates online learning. (+S) indicates the use of static image datasets during the network training.}
		\vspace{2mm}
		\small
		\begin{tabular}{lP{0.58cm}P{0.58cm}P{0.58cm}P{0.58cm}P{0.58cm}}
			\toprule
			Method &OL &fps &$\mathcal{G}_\mathcal{M}$ &$\mathcal{J}_\mathcal{M}$ &$\mathcal{F}_\mathcal{M}$\\
			\midrule
			FEELVOS~(+S)~\cite{FEELVOS} & &2.22 &69.1 &65.9 &72.3\\
			GC~(+S)~\cite{GC} & &25.0 &71.4 &69.3 &73.5\\
			STM~(+S)~\cite{STM} & &6.25 &71.6 &69.2 &74.0\\
			LWL~(+S)~\cite{LWL} &\checkmark &14.0 &74.3 &72.2 &76.3\\
			AFB-URR~(+S)~\cite{AFB-URR} & &4.00 &74.6 &73.0 &76.1\\
			CFBI~(+S)~\cite{CFBI} & &5.56 &74.9 &72.1 &77.7\\
			RMNet~(+S)~\cite{RMNet} & &11.9 &75.0 &72.8 &77.2\\
			LCM~(+S)~\cite{LCM} & &8.47 &75.2 &73.1 &77.2\\
			PReMVOS~(+S)~\cite{PReMVOS} &\checkmark &0.03 &\textbf{77.8} &\textbf{73.9} &\textbf{81.7}\\
			\midrule
			STM~\cite{STM} & &6.25 &43.0 &38.1 &47.9\\
			FAVOS~\cite{FAVOS} & &0.56 &58.2 &54.6 &61.8\\
			VideoMatch~\cite{VideoMatch} & &3.13 &62.4 &56.5 &68.2\\
			AGSS-VOS~\cite{AGSS-VOS} & &10.0 &66.6 &63.4 &69.8\\
			FRTM~\cite{FRTM} &\checkmark &21.9 &68.8 &66.4 &71.2\\
			\midrule
			\textbf{BMVOS} & &\textbf{45.9} &\textbf{72.7} &\textbf{70.7} &\textbf{74.7}\\
			\bottomrule
		\end{tabular}
		\label{Table:DAVIS17_val}
	\end{table}

	\begin{table}
		\centering 
		\caption{Quantitative evaluation on the DAVIS 2017 test-dev set. OL indicates online learning. (+S) indicates the use of static image datasets during the network training.}
		\vspace{2mm}
		\small
		\begin{tabular}{lP{0.58cm}P{0.58cm}P{0.58cm}P{0.58cm}P{0.58cm}}
			\toprule
			Method &OL &fps &$\mathcal{G}_\mathcal{M}$ &$\mathcal{J}_\mathcal{M}$ &$\mathcal{F}_\mathcal{M}$\\
			\midrule
			RGMP~(+S)~\cite{RGMP} & &7.69 &52.9 &51.3 &54.4\\
			FEELVOS~(+S)~\cite{FEELVOS} & &2.22 &54.4 &51.2 &57.5\\
			RANet~(+S)~\cite{RANet} & &30.3 &55.3 &53.4 &57.2\\
			OSVOS-S~(+S)~\cite{OSVOS-S} &\checkmark &0.22 &57.5 &52.9 &62.1\\
			CNN-MRF~(+S)~\cite{CNN-MRF} &\checkmark &0.03 &67.5 &64.5 &70.5\\
			DyeNet~(+S)~\cite{DyeNet} &\checkmark &0.43 &68.2 &65.8 &70.5\\
			PReMVOS~(+S)~\cite{PReMVOS} &\checkmark &0.03 &\textbf{71.6} &\textbf{67.5} &\textbf{75.7}\\
			\midrule
			OSMN~\cite{OSMN} & &7.14 &41.3 &37.7 &44.9\\
			FAVOS~\cite{FAVOS} & &0.56 &43.6 &42.9 &44.2\\
			AGSS-VOS~\cite{AGSS-VOS} & &10.0 &54.3 &51.5 &57.1\\
			\midrule
			\textbf{BMVOS} & &\textbf{45.9} &\textbf{62.7} &\textbf{60.7} &\textbf{64.7}\\
			\bottomrule
		\end{tabular}
		\label{Table:DAVIS17_test}
	\end{table}

	\noindent\textbf{DAVIS 2016.} In Table~\ref{Table:DAVIS16_val}, we compare our method with state-of-the-art methods on the DAVIS 2016 validation set. The best performance is obtained by KMN~\cite{KMN}, with a $\mathcal{G}$ score of 87.6\%. However, its inference speed is not satisfactory for real-world systems and also requires extensive static image datasets for network training. Of the methods trained without static image datasets, we outperform all previous approaches, with a $\mathcal{G}$ score of 82.2\%.

	\begin{figure*}[t]
		\centering
		\includegraphics[width=1.0\linewidth]{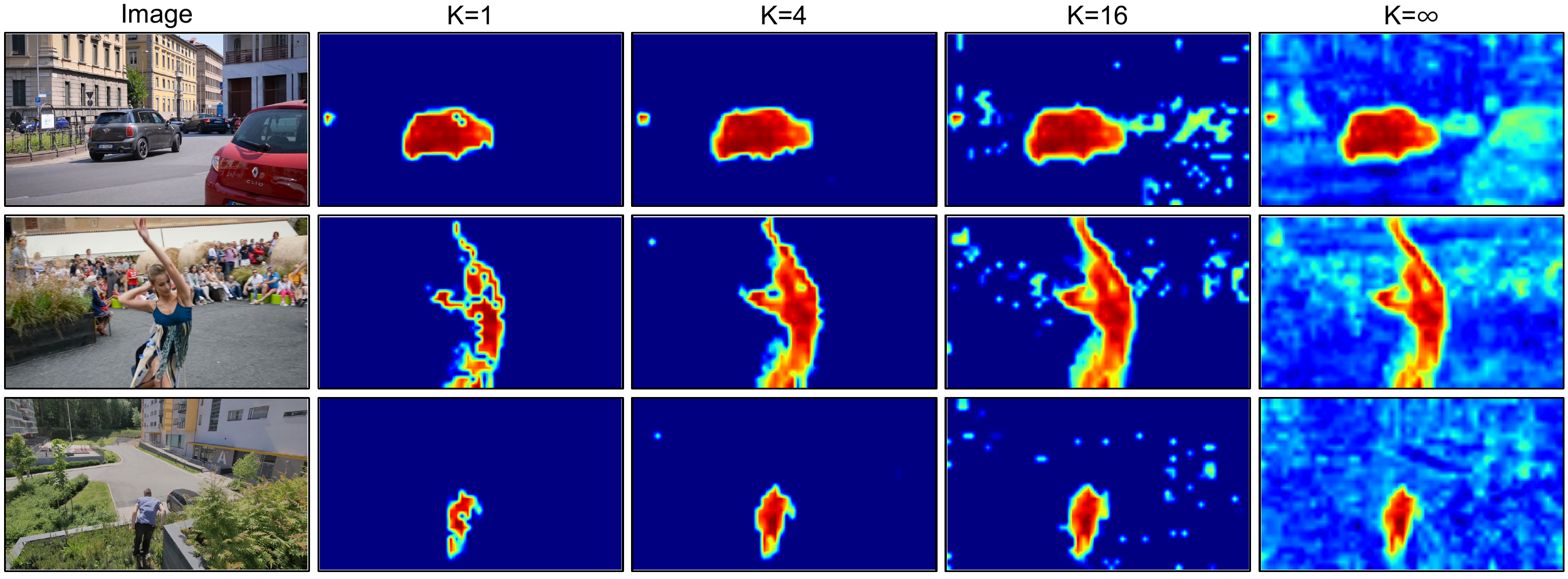}
		\caption{Comparison of similarity matching score maps with various K values. For a clear visualization, we linearly normalize the values in each score map to be in the same range.}
		\label{figure5}
	\end{figure*}

	\vspace{1mm} 
	\noindent\textbf{DAVIS 2017.} Comparison on the DAVIS 2017 validation set and test-dev set can be found at Table~\ref{Table:DAVIS17_val} and Table~\ref{Table:DAVIS17_test}, respectively. As most state-of-the-art methods~\cite{STM, GC, KMN, CFBI, RMNet, LCM} do not report their inference speed on these datasets, we assume each video sequence contains a single target object. On both datasets, our method shows a remarkable performance even maintaining fast inference speed and without using static image training datasets. Compared to the methods trained using the same amount of data, it outperforms all existing methods by a large margin, demonstrating its exceptional generalization ability.

	\begin{table}
		\centering 
		\caption{Quantitative evaluation on the YouTube-VOS 2018 validation set. (+S) indicates the use of static image datasets during the network training.}
		\vspace{2mm}
		\small
		\begin{tabular}{p{2.2cm}P{0.52cm}P{0.52cm}P{0.52cm}P{0.52cm}P{0.52cm}P{0.52cm}}
			\toprule
			Method &fps &$\mathcal{G}_\mathcal{M}$ &$\mathcal{J}_\mathcal{S}$ &$\mathcal{J}_\mathcal{U}$ &$\mathcal{F}_\mathcal{S}$ &$\mathcal{F}_\mathcal{U}$\\
			\midrule
			SAT~(+S)~\cite{SAT} &\textbf{39.0} &63.6 &67.1 &55.3 &70.2 &61.7\\
			GC~(+S)~\cite{GC} &- &73.2 &72.6 &68.9 &75.6 &75.7\\
			STM~(+S)~\cite{STM} &- &79.4 &79.7 &72.8 &84.2 &80.9\\
			CFBI~(+S)~\cite{CFBI} &- &81.4 &81.1 &75.3 &85.8 &83.4\\
			KMN~(+S)~\cite{KMN} &- &81.4 &81.4 &75.3 &85.6 &83.3\\
			LWL~(+S)~\cite{LWL} &- &81.5 &80.4 &\textbf{76.4} &84.9 &\textbf{84.4}\\
		    LCM~(+S)~\cite{LCM} &- &\textbf{82.0} &\textbf{82.2} &75.7 &\textbf{86.7}&83.4\\
			\midrule
			RVOS~\cite{RVOS} &22.7 &56.8 &63.6 &45.5 &67.2 &51.0\\
			CapsuleVOS~\cite{CapsuleVOS} &13.5 &62.3 &67.3 &53.7 &68.1 &59.9\\
			S2S~\cite{S2S} &- &64.4 &71.0 &55.5 &70.0 &61.2\\
			STM~\cite{STM} &- &68.2 &- &- &- &-\\
			AGSS-VOS~\cite{AGSS-VOS} &12.5 &71.3 &71.3 &65.5 &76.2 &73.1\\
			FRTM~\cite{FRTM} &- &72.1 &72.3 &65.9 &76.2 &74.1\\
			\midrule
			\textbf{BMVOS} &28.0 &\textbf{73.9} &\textbf{73.5} &\textbf{68.5} &\textbf{77.4} &\textbf{76.0}\\
			\bottomrule
		\end{tabular}
		\label{Table:YTVOS}
	\end{table}

	\vspace{1mm} 
	\noindent\textbf{YouTube-VOS 2018.} In Table~\ref{Table:YTVOS}, our method is compared to state-of-the-art methods on the YouTube-VOS 2018 validation set. As can be seen through STM~\cite{STM}, using static image datasets can dramatically improve the performance of the network. Even without them, our method achieves a competitive accuracy compared to the state-of-the-art methods with quite a fast inference speed. We use the original resolution if the number of foreground pixels is less than 1,000 so as to capture the small objects, while otherwise we resize the videos to 480p.

	\subsection{Ablation Study}

	\begin{figure*}[t]
		\centering
		\includegraphics[width=1.0\linewidth]{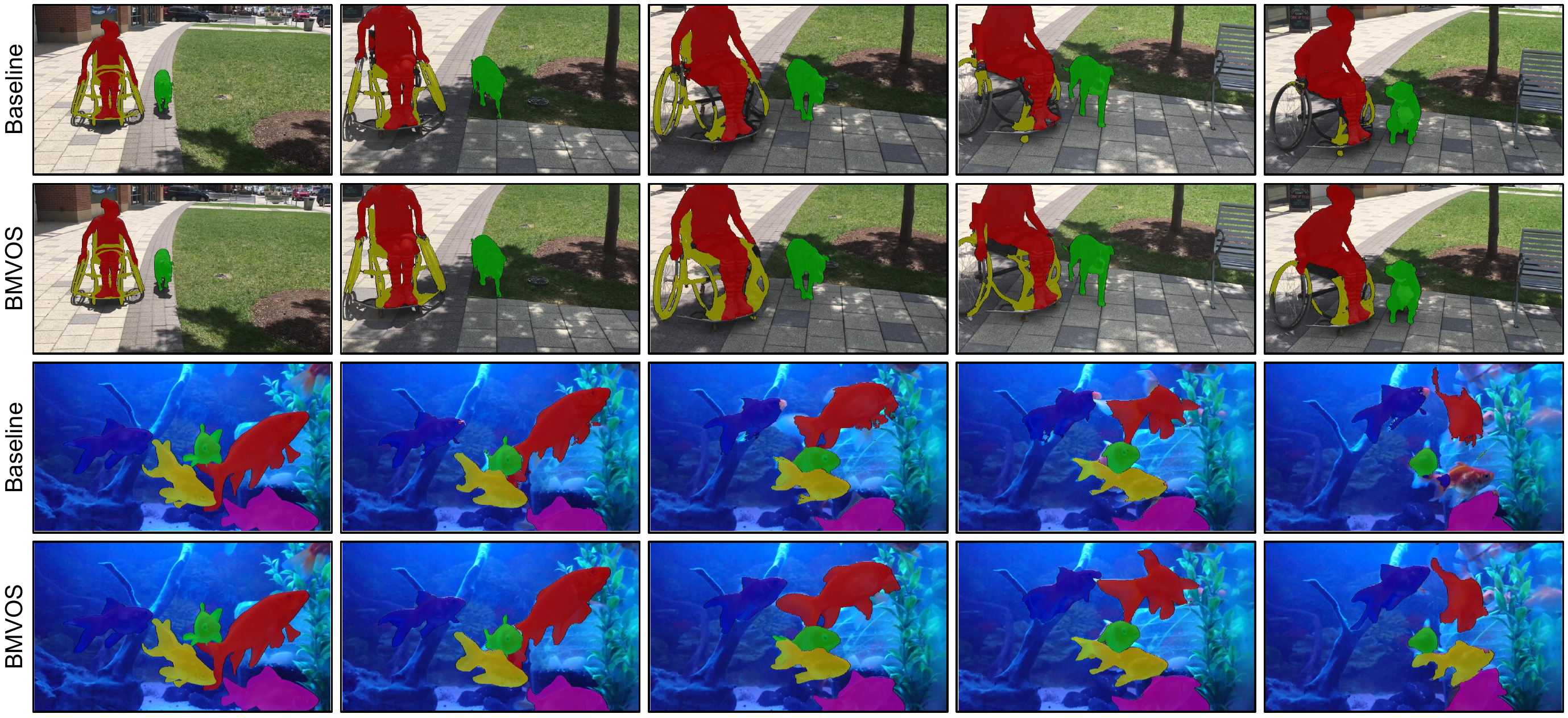}
		\caption{Qualitative comparison of our baseline model and BMVOS on the DAVIS 2017 validation set and test-dev set.}
		\label{figure6}
	\end{figure*}

	\label{ablation}
	\noindent\textbf{Bijective matching strictness.} To obtain the flexibility and reliability of prediction at the same time, we exploit surjective matching for global matching and bijective matching for local matching. Unlike conventional surjective matching, our proposed bijective matching performs a reference-wise top K operation to ensure the stability of prediction. Based on the K value, the strictness of matching can be flexibly adjusted. For example, if K is set to a large value, the matching will be performed similar to surjective matching, not strictly limiting the amount of each reference frame pixel being referenced. On the other hand, if K is set to a small value, the matching will be performed strictly, i.e., information will be transferred between two pixels if and only if they are the sure matches for each other. To prove this property, we perform extensive ablation studies on the K value. In Figure~\ref{figure5}, we visualize the similarity matching score maps with various K values. If K is set to infinity, the matching mechanism is equivalent to a surjective matching as there is no limitation to the number of reference frame pixels being referenced. As K increases, the matching is performed in a more relaxed manner, so the background distractors that are similar to a target object get higher scores. Whereas if K is set too small, the matching is performed overly strictly and may harm the smoothness and completeness of the segmentation mask. Therefore, we carefully select the K value to strike a balance between strictness and smoothness. As can be seen in Table~\ref{Table:ablation}, the optimal performance is obtained when K is set to infinity for global matching and 4 for local matching. It is notable that the global matching achieves the optimal performance when K is set to infinity as two frames are not interdependent enough. Adopting bijective matching for local matching brings about a 2\% accuracy improvement on $\mathcal{G_M}$.

	\begin{table}
		\centering 
		\caption{Ablation study on bijective matching strictness and mask propagation. K1 and K2 are the K values for global matching and local matching, respectively. ME indicates the use of our proposed mask embedding module and L is the number of frames used for mask propagation. All model versions are trained on the DAVIS 2017 training set and tested on the DAVIS 2017 validation set.}
		\vspace{2mm}
		\small
		\begin{tabular}{c|P{0.5cm}|P{0.5cm}|P{0.5cm}|P{0.5cm}|P{0.58cm}P{0.58cm}P{0.58cm}}
			\toprule
			Version &K1 &K2 &ME &L &$\mathcal{G}_\mathcal{M}$ &$\mathcal{J}_\mathcal{M}$ &$\mathcal{F}_\mathcal{M}$\\
			\midrule
			Baseline &$\infty$ & $\infty$ &- &1 &67.4 &65.2 &69.6\\			
			\midrule			
			\Romannum{1} &$\infty$ &1 &- &1 &68.0 &66.1 &69.8\\
			\Romannum{2} &$\infty$ &2 &- &1 &69.1 &67.2 &71.1\\				
			\Romannum{3} &$\infty$ &4 &- &1 &69.3 &67.2 &71.3\\
			\Romannum{4} &$\infty$ &8 &- &1 &68.5 &66.4 &70.6\\
			\Romannum{5} &$\infty$ &16 &- &1 &68.1 &66.0 &70.3\\
			\midrule
			\Romannum{6} &1 & 4 &- &1 &65.8 &64.4 &67.1\\
			\Romannum{7} &2 & 4 &- &1 &66.8 &65.5 &68.1\\
			\Romannum{8} &4 & 4 &- &1 &67.5 &66.0 &69.1\\
			\Romannum{9} &8 & 4 &- &1 &68.4 &66.4 &70.3\\
			\Romannum{10} &16 & 4 &- &1 &68.5 &66.3 &70.7\\
			\midrule
			\Romannum{11} &$\infty$ &4 &\checkmark &1 &71.2 &69.3 &73.1\\
			\Romannum{12} &$\infty$ &4 &\checkmark &2 &72.0 &69.7 &74.3\\
			\Romannum{13} &$\infty$ &4 &\checkmark &3 &72.7 &70.7 &74.7\\
			\Romannum{14} &$\infty$ &4 &\checkmark &4 &72.3 &70.5 &74.2\\
			\Romannum{15} &$\infty$ &4 &\checkmark &5 &72.3 &70.4 &74.2\\
			\bottomrule
		\end{tabular}
		\label{Table:ablation}
	\end{table}

	\vspace{1mm}
	\noindent\textbf{Mask propagation.} To bolster the appearance information, the position information of a target object is widely used in VOS. A popular approach is just feeding a downsampled previous frame mask into a decoder for mask propagation. To better utilize the position information of the target object, we propose a mask embedding module that extracts the features from multiple historic frames with spatial coordinate information. In Table~\ref{Table:ablation}, we compare various model versions with and without mask embedding module, using different number of frames for mask propagation. The best performance is obtained when using three historic frames with mask embedding module, with a $\mathcal{G}$ score of 72.7\%. Compared to the model without mask embedding module, it dramatically boosts the $\mathcal{G}$ score by 3.4\%.

	\vspace{1mm}
	\noindent\textbf{Qualitative comparison to baseline.} We qualitatively compare BMVOS to our baseline model in Figure~\ref{figure6}. Due to its bijective local matching and mask embedding module, BMVOS is more effective for capturing temporal consistency of a video compared to the baseline model.

	\section{Conclusion}
	Existing VOS methods employ feature matching to capture the appearance information of a target object. However, due to its surjective nature, it is susceptible to background distractions that may cause critical errors. To mitigate this concern, we introduce a bijective matching mechanism to better transfer information from the reference frame to the query frame. In addition, we improve existing mask propagation method by proposing a mask embedding module. On public benchmark datasets, our proposed approach demonstrates its effectiveness while being simple and straightforward. We hope our method will be widely used for future VOS research and the fields in computer vision.

	\vspace*{\fill}
	\noindent\footnotesize\textbf{Acknowledgement.} This research was supported by Multi-Ministry Collaborative R\&D Program(R\&D program for complex cognitive technology) through the National Research Foundation of Korea(NRF) funded by MSIT, MOTIE, KNPA(NRF-2018M3E3A1057289).

	{\small
		\bibliographystyle{ieee}
		\bibliography{bib}}

\begin{thebibliography}{10}\itemsep=-1pt

\bibitem{CNN-MRF}
Linchao Bao, Baoyuan Wu, and Wei Liu.
\newblock Cnn in mrf: Video object segmentation via inference in a cnn-based
  higher-order spatio-temporal mrf.
\newblock In {\em Proceedings of the IEEE conference on computer vision and
  pattern recognition}, pages 5977--5986, 2018.

\bibitem{LWL}
Goutam Bhat, Felix~J{\"a}remo Lawin, Martin Danelljan, Andreas Robinson,
  Michael Felsberg, Luc Van~Gool, and Radu Timofte.
\newblock Learning what to learn for video object segmentation.
\newblock In {\em European Conference on Computer Vision}, pages 777--794.
  Springer, 2020.

\bibitem{SAT}
Xi Chen, Zuoxin Li, Ye Yuan, Gang Yu, Jianxin Shen, and Donglian Qi.
\newblock State-aware tracker for real-time video object segmentation.
\newblock In {\em Proceedings of the IEEE/CVF Conference on Computer Vision and
  Pattern Recognition}, pages 9384--9393, 2020.

\bibitem{FAVOS}
Jingchun Cheng, Yi-Hsuan Tsai, Wei-Chih Hung, Shengjin Wang, and Ming-Hsuan
  Yang.
\newblock Fast and accurate online video object segmentation via tracking
  parts.
\newblock In {\em Proceedings of the IEEE conference on computer vision and
  pattern recognition}, pages 7415--7424, 2018.

\bibitem{MSRA10K}
Ming-Ming Cheng, Niloy~J Mitra, Xiaolei Huang, Philip~HS Torr, and Shi-Min Hu.
\newblock Global contrast based salient region detection.
\newblock {\em IEEE transactions on pattern analysis and machine intelligence},
  37(3):569--582, 2014.

\bibitem{CRVOS}
Suhwan Cho, MyeongAh Cho, Tae-young Chung, Heansung Lee, and Sangyoun Lee.
\newblock Crvos: Clue refining network for video object segmentation.
\newblock In {\em 2020 IEEE International Conference on Image Processing
  (ICIP)}, pages 2301--2305. IEEE, 2020.

\bibitem{CapsuleVOS}
Kevin Duarte, Yogesh~S Rawat, and Mubarak Shah.
\newblock Capsulevos: Semi-supervised video object segmentation using capsule
  routing.
\newblock In {\em Proceedings of the IEEE International Conference on Computer
  Vision}, pages 8480--8489, 2019.

\bibitem{SSTVOS}
Brendan Duke, Abdalla Ahmed, Christian Wolf, Parham Aarabi, and Graham~W
  Taylor.
\newblock Sstvos: Sparse spatiotemporal transformers for video object
  segmentation.
\newblock In {\em Proceedings of the IEEE/CVF Conference on Computer Vision and
  Pattern Recognition}, pages 5912--5921, 2021.

\bibitem{PASCAL-VOC}
Mark Everingham, Luc Van~Gool, Christopher~KI Williams, John Winn, and Andrew
  Zisserman.
\newblock The pascal visual object classes (voc) challenge.
\newblock {\em International journal of computer vision}, 88(2):303--338, 2010.

\bibitem{GIEL}
Wenbin Ge, Xiankai Lu, and Jianbing Shen.
\newblock Video object segmentation using global and instance embedding
  learning.
\newblock In {\em Proceedings of the IEEE/CVF Conference on Computer Vision and
  Pattern Recognition}, pages 16836--16845, 2021.

\bibitem{LCM}
Li Hu, Peng Zhang, Bang Zhang, Pan Pan, Yinghui Xu, and Rong Jin.
\newblock Learning position and target consistency for memory-based video
  object segmentation.
\newblock In {\em Proceedings of the IEEE/CVF Conference on Computer Vision and
  Pattern Recognition}, pages 4144--4154, 2021.

\bibitem{VideoMatch}
Yuan-Ting Hu, Jia-Bin Huang, and Alexander~G Schwing.
\newblock Videomatch: Matching based video object segmentation.
\newblock In {\em Proceedings of the European Conference on Computer Vision
  (ECCV)}, pages 54--70, 2018.

\bibitem{densenet}
Gao Huang, Zhuang Liu, Laurens Van Der~Maaten, and Kilian~Q Weinberger.
\newblock Densely connected convolutional networks.
\newblock In {\em Proceedings of the IEEE conference on computer vision and
  pattern recognition}, pages 4700--4708, 2017.

\bibitem{batchnorm}
Sergey Ioffe and Christian Szegedy.
\newblock Batch normalization: Accelerating deep network training by reducing
  internal covariate shift.
\newblock {\em arXiv preprint arXiv:1502.03167}, 2015.

\bibitem{A-GAME}
Joakim Johnander, Martin Danelljan, Emil Brissman, Fahad~Shahbaz Khan, and
  Michael Felsberg.
\newblock A generative appearance model for end-to-end video object
  segmentation.
\newblock In {\em Proceedings of the IEEE Conference on Computer Vision and
  Pattern Recognition}, pages 8953--8962, 2019.

\bibitem{adam}
Diederik~P Kingma and Jimmy Ba.
\newblock Adam: A method for stochastic optimization.
\newblock {\em arXiv preprint arXiv:1412.6980}, 2014.

\bibitem{imagenet}
Alex Krizhevsky, Ilya Sutskever, and Geoffrey~E Hinton.
\newblock Imagenet classification with deep convolutional neural networks.
\newblock {\em Communications of the ACM}, 60(6):84--90, 2017.

\bibitem{DyeNet}
Xiaoxiao Li and Chen Change~Loy.
\newblock Video object segmentation with joint re-identification and
  attention-aware mask propagation.
\newblock In {\em Proceedings of the European Conference on Computer Vision
  (ECCV)}, pages 90--105, 2018.

\bibitem{PASCAL-S}
Yin Li, Xiaodi Hou, Christof Koch, James~M Rehg, and Alan~L Yuille.
\newblock The secrets of salient object segmentation.
\newblock In {\em Proceedings of the IEEE Conference on Computer Vision and
  Pattern Recognition}, pages 280--287, 2014.

\bibitem{GC}
Yu Li, Zhuoran Shen, and Ying Shan.
\newblock Fast video object segmentation using the global context module.
\newblock {\em arXiv preprint arXiv:2001.11243}, 2020.

\bibitem{AFB-URR}
Yongqing Liang, Xin Li, Navid Jafari, and Qin Chen.
\newblock Video object segmentation with adaptive feature bank and
  uncertain-region refinement.
\newblock {\em arXiv preprint arXiv:2010.07958}, 2020.

\bibitem{AGSS-VOS}
Huaijia Lin, Xiaojuan Qi, and Jiaya Jia.
\newblock Agss-vos: Attention guided single-shot video object segmentation.
\newblock In {\em Proceedings of the IEEE International Conference on Computer
  Vision}, pages 3949--3957, 2019.

\bibitem{COCO}
Tsung-Yi Lin, Michael Maire, Serge Belongie, James Hays, Pietro Perona, Deva
  Ramanan, Piotr Doll{\'a}r, and C~Lawrence Zitnick.
\newblock Microsoft coco: Common objects in context.
\newblock In {\em European conference on computer vision}, pages 740--755.
  Springer, 2014.

\bibitem{coord}
Rosanne Liu, Joel Lehman, Piero Molino, Felipe~Petroski Such, Eric Frank, Alex
  Sergeev, and Jason Yosinski.
\newblock An intriguing failing of convolutional neural networks and the
  coordconv solution.
\newblock In {\em Advances in Neural Information Processing Systems}, pages
  9605--9616, 2018.

\bibitem{EGMN}
Xiankai Lu, Wenguan Wang, Martin Danelljan, Tianfei Zhou, Jianbing Shen, and
  Luc Van~Gool.
\newblock Video object segmentation with episodic graph memory networks.
\newblock In {\em Computer Vision--ECCV 2020: 16th European Conference,
  Glasgow, UK, August 23--28, 2020, Proceedings, Part III 16}, pages 661--679.
  Springer, 2020.

\bibitem{PReMVOS}
Jonathon Luiten, Paul Voigtlaender, and Bastian Leibe.
\newblock Premvos: Proposal-generation, refinement and merging for video object
  segmentation.
\newblock In {\em Asian Conference on Computer Vision}, pages 565--580.
  Springer, 2018.

\bibitem{OSVOS-S}
K-K Maninis, Sergi Caelles, Yuhua Chen, Jordi Pont-Tuset, Laura Leal-Taix{\'e},
  Daniel Cremers, and Luc Van~Gool.
\newblock Video object segmentation without temporal information.
\newblock {\em IEEE transactions on pattern analysis and machine intelligence},
  41(6):1515--1530, 2018.

\bibitem{relu}
Vinod Nair and Geoffrey~E Hinton.
\newblock Rectified linear units improve restricted boltzmann machines.
\newblock In {\em Icml}, 2010.

\bibitem{STM}
Seoung~Wug Oh, Joon-Young Lee, Ning Xu, and Seon~Joo Kim.
\newblock Video object segmentation using space-time memory networks.
\newblock In {\em Proceedings of the IEEE International Conference on Computer
  Vision}, pages 9226--9235, 2019.

\bibitem{DAVIS2016}
Federico Perazzi, Jordi Pont-Tuset, Brian McWilliams, Luc Van~Gool, Markus
  Gross, and Alexander Sorkine-Hornung.
\newblock A benchmark dataset and evaluation methodology for video object
  segmentation.
\newblock In {\em Proceedings of the IEEE Conference on Computer Vision and
  Pattern Recognition}, pages 724--732, 2016.

\bibitem{DAVIS2017}
Jordi Pont-Tuset, Federico Perazzi, Sergi Caelles, Pablo Arbel\'aez, Alexander
  Sorkine-Hornung, and Luc {Van Gool}.
\newblock The 2017 davis challenge on video object segmentation.
\newblock {\em arXiv:1704.00675}, 2017.

\bibitem{FRTM}
Andreas Robinson, Felix~Jaremo Lawin, Martin Danelljan, Fahad~Shahbaz Khan, and
  Michael Felsberg.
\newblock Learning fast and robust target models for video object segmentation.
\newblock In {\em Proceedings of the IEEE/CVF Conference on Computer Vision and
  Pattern Recognition}, pages 7406--7415, 2020.

\bibitem{KMN}
Hongje Seong, Junhyuk Hyun, and Euntai Kim.
\newblock Kernelized memory network for video object segmentation.
\newblock In {\em European Conference on Computer Vision}, pages 629--645.
  Springer, 2020.

\bibitem{ECSSD}
Jianping Shi, Qiong Yan, Li Xu, and Jiaya Jia.
\newblock Hierarchical image saliency detection on extended cssd.
\newblock {\em IEEE transactions on pattern analysis and machine intelligence},
  38(4):717--729, 2015.

\bibitem{PLM}
Jae Shin~Yoon, Francois Rameau, Junsik Kim, Seokju Lee, Seunghak Shin, and In
  So~Kweon.
\newblock Pixel-level matching for video object segmentation using
  convolutional neural networks.
\newblock In {\em Proceedings of the IEEE international conference on computer
  vision}, pages 2167--2176, 2017.

\bibitem{RVOS}
Carles Ventura, Miriam Bellver, Andreu Girbau, Amaia Salvador, Ferran Marques,
  and Xavier Giro-i Nieto.
\newblock Rvos: End-to-end recurrent network for video object segmentation.
\newblock In {\em Proceedings of the IEEE/CVF Conference on Computer Vision and
  Pattern Recognition}, pages 5277--5286, 2019.

\bibitem{FEELVOS}
Paul Voigtlaender, Yuning Chai, Florian Schroff, Hartwig Adam, Bastian Leibe,
  and Liang-Chieh Chen.
\newblock Feelvos: Fast end-to-end embedding learning for video object
  segmentation.
\newblock In {\em Proceedings of the IEEE Conference on Computer Vision and
  Pattern Recognition}, pages 9481--9490, 2019.

\bibitem{RANet}
Ziqin Wang, Jun Xu, Li Liu, Fan Zhu, and Ling Shao.
\newblock Ranet: Ranking attention network for fast video object segmentation.
\newblock In {\em Proceedings of the IEEE international conference on computer
  vision}, pages 3978--3987, 2019.

\bibitem{RGMP}
Seoung Wug~Oh, Joon-Young Lee, Kalyan Sunkavalli, and Seon Joo~Kim.
\newblock Fast video object segmentation by reference-guided mask propagation.
\newblock In {\em Proceedings of the IEEE conference on computer vision and
  pattern recognition}, pages 7376--7385, 2018.

\bibitem{RMNet}
Haozhe Xie, Hongxun Yao, Shangchen Zhou, Shengping Zhang, and Wenxiu Sun.
\newblock Efficient regional memory network for video object segmentation.
\newblock In {\em Proceedings of the IEEE/CVF Conference on Computer Vision and
  Pattern Recognition}, pages 1286--1295, 2021.

\bibitem{S2S}
Ning Xu, Linjie Yang, Yuchen Fan, Jianchao Yang, Dingcheng Yue, Yuchen Liang,
  Brian Price, Scott Cohen, and Thomas Huang.
\newblock Youtube-vos: Sequence-to-sequence video object segmentation.
\newblock In {\em Proceedings of the European conference on computer vision
  (ECCV)}, pages 585--601, 2018.

\bibitem{YTVOS}
Ning Xu, Linjie Yang, Yuchen Fan, Dingcheng Yue, Yuchen Liang, Jianchao Yang,
  and Thomas Huang.
\newblock Youtube-vos: A large-scale video object segmentation benchmark.
\newblock {\em arXiv preprint arXiv:1809.03327}, 2018.

\bibitem{OSMN}
Linjie Yang, Yanran Wang, Xuehan Xiong, Jianchao Yang, and Aggelos~K
  Katsaggelos.
\newblock Efficient video object segmentation via network modulation.
\newblock In {\em Proceedings of the IEEE Conference on Computer Vision and
  Pattern Recognition}, pages 6499--6507, 2018.

\bibitem{CFBI}
Zongxin Yang, Yunchao Wei, and Yi Yang.
\newblock Collaborative video object segmentation by foreground-background
  integration.
\newblock {\em arXiv preprint arXiv:2003.08333}, 2020.

\bibitem{deconv}
Matthew~D Zeiler, Graham~W Taylor, and Rob Fergus.
\newblock Adaptive deconvolutional networks for mid and high level feature
  learning.
\newblock In {\em 2011 International Conference on Computer Vision}, pages
  2018--2025. IEEE, 2011.

\bibitem{DTN}
Lu Zhang, Zhe Lin, Jianming Zhang, Huchuan Lu, and You He.
\newblock Fast video object segmentation via dynamic targeting network.
\newblock In {\em Proceedings of the IEEE/CVF International Conference on
  Computer Vision}, pages 5582--5591, 2019.

\bibitem{TVOS}
Yizhuo Zhang, Zhirong Wu, Houwen Peng, and Stephen Lin.
\newblock A transductive approach for video object segmentation.
\newblock In {\em Proceedings of the IEEE/CVF Conference on Computer Vision and
  Pattern Recognition}, pages 6949--6958, 2020.

\end{thebibliography}
	
\end{document}